\documentclass[numbers]{article}

\usepackage[final]{nips_2017}

\usepackage[utf8]{inputenc} 
\usepackage[T1]{fontenc}    
\usepackage{hyperref}       
\usepackage{url}            
\usepackage{booktabs}       
\usepackage{amsfonts}       
\usepackage{amsmath}        
\usepackage{nicefrac}       
\usepackage{microtype}      
\usepackage{bigints}        
\usepackage{graphicx}       
\usepackage{subfig}         
\usepackage{hyperref}       
\usepackage{color}

\usepackage{natbib}

\usepackage{todonotes}

\newcommand*\samethanks[1][\value{footnote}]{\footnotemark[#1]}

\title{Alpha-Divergences in Variational Dropout}

\author{
   Bogdan Mazoure \thanks{These two authors contributed equally} \\
   Department of Mathematics and Statistics\\
   McGill University\\
   Montreal, QC H3A 0G4 \\
   \texttt{bogdan.mazoure@mail.mcgill.ca} \\  
   \And
  Riashat Islam \samethanks\\
  Department of Computer Science\\
  McGill University\\
  Montreal, QC H3A 0G4 \\
  \texttt{riashat.islam@mail.mcgill.ca} \\
}

\begin{document}

\maketitle

\begin{abstract}

We investigate the use of alternative divergences to Kullback-Leibler (KL) in variational inference(VI), based on the Variational Dropout \cite{kingma2015}. Stochastic gradient variational Bayes (SGVB) \cite{aevb} is a general framework for estimating the evidence lower bound (ELBO) in Variational Bayes. In this work, we extend the SGVB estimator with using Alpha-Divergences, which are alternative to divergences to VI' KL objective. The Gaussian dropout can be seen as a local reparametrization trick of the SGVB objective. We extend the Variational Dropout to use alpha divergences for variational inference. Our results compare $\alpha$-divergence variational dropout with standard variational dropout with correlated and uncorrelated weight noise. We show that the $\alpha$-divergence with $\alpha \rightarrow 1$ (or KL divergence) is still a good measure for use in variational inference, in spite of the efficient use of Alpha-divergences for Dropout VI \cite{Li17}. $\alpha \rightarrow 1$ can yield the lowest training error, and optimizes a good lower bound for the evidence lower bound (ELBO) among all values of the parameter $\alpha \in [0,\infty)$. 
\end{abstract}

\section{Introduction}

Deep Neural Networks have achieved tremendous success in recent years, due to the ability of these models to achieve state-of-the-art performances \cite{krizhevsky2012,lecun2015,Szegedy16}. These architectures contain a large number of parameters, and several regularization techniques have been proposed for these models to avoid overfitting \cite{srivastava2014}, which can impact the network's ability to generalize. Regularization techniques have been developed to constrain the number of active parameters in a large model. For instance, dropout \cite{srivastava2014} adds multiplicative noise to the input matrix while batch normalization \cite{ioffe2015batch} reduces the internal covariate shift by standardizing the weights to zero mean unit variance over each dimension. In addition to these new regularization techniques, standard methods like ridge and lasso can be used. However, as in \cite{kingma2015}, a Bayesian treatment of dropout through the minimization of the Kullback-Leibler divergence between the weight priors and and approximate posterior also leads to regularization.

Bayesian deep learning has also recently gained significant popularity, due to the ability of Bayesian neural networks to model uncertainty. This has increased the scope of applications in deep learning, ranging from classification and regression \cite{gal2016}, active learning \cite{DBLP:conf/icml/GalIG17} and tasks to detect adversarial examples \cite{DBLP:journals/corr/abs-1710-04759}. The success of Bayesian neural networks is largely due to being able to apply approximate inference techniques such as variational inference \cite{DBLP:journals/ml/JordanGJS99}. Many different approximations to Bayesian neural networks using variational inference has been proposed \cite{PBP}, \cite{gal2016, DBLP:journals/corr/abs-1710-04759, DBLP:journals/corr/BlundellCKW15} in recent years. \cite{kingma2015} also showed that the Gaussian version of dropout has a neat Bayesian connection and can be used to represent model uncertainty, although counter arguments show that it may not be a Bayesian approach \cite{hiron}. Variational inference, however, using the KL divergence lies at the core of almost all approaches to Bayesian deep learning.

Dropout \cite{srivastava2014} is a regularization technique which adds multiplicative random noise to each input matrix of a given neural layer. The most common version of dropout uses the binary, or Bernoulli distribution. Gaussian dropout as an approximation to the binary noise was later proposed (\cite{wang2013}). Recently, \cite{gal2016} showed that dropout can be considered as an approximation to variational inference in neural networks. Alternatively, \cite{kingma2015} introduced the variational dropout which generalizes the continuous dropout technique over independent and dependent weight noise. Due to it's convenient mathematical properties and close connection to entropy, the KL divergence is very widely used in Bayesian techniques and variational inference. Even though the KL divergence can underestimate uncertainty significantly, it has been widely used for approximate inference in Bayesian neural networks. Recently, \cite{Li17} showed that $\alpha$-divergence can be used instead of the KL divergence for approximate inference in dropout variational inference \cite{gal2016}.

In this work, we examine the significance of using $\alpha$-divergence in variational dropout originally proposed in \cite{kingma2015}. Similarly to \cite{li2016renyi}, we derive a new variational lower bound by generalizing the KL divergence to $\alpha$-divergences. We propose the use of $\alpha$-divergence that can be used in variational dropout. We justify that $\alpha$-divergence is a general case of the Kullback-Leibler, Hellinger, total variation, maximum probability ratio and $\chi^2$ divergences. Our experiments on the MNIST dataset shows that the variational dropout with correlated noise achieves the best performance when the $\alpha$-divergence tends to the KL divergence.

\section{Background}

\subsection{Dropout}

For a dense neural network layer $h$, the output of $h$ after application of dropout will have the form:

\begin{equation}
O_h=g\big(\big(X_h \circ \varepsilon\big)W_h\big),\;\; \varepsilon \sim p(\varepsilon),
\label{eq:dropout}
\end{equation}
where $X_h \in \mathbb{R}^M \times \mathbb{R}^K$ is the feature matrix obtained from the previous layer $h-1$, $W_h \in \mathbb{R}^K \times \mathbb{R}^L$ is the weight matrix for the layer, $\varepsilon \in \mathbb{R}^M \times \mathbb{R}^K $ is the dropout noise matrix, $g(\cdot)$ is the activation function, $\circ$ is the entrywise Haddamard product and $O_n \in \mathbb{R}^M \times \mathbb{R}^L$ is the output matrix of layer $h$ after the activation function. The hyperparameter in the dropout technique is considered to be the \textit{dropout rate}, $p$. It is analogous to the regularization intensity $\lambda$ for $L_p$ norm regularization and will control the magnitude of the dropout effect. The dropout rate is constrained to be a probability between $0$ and $1$ inclusively. In the case of binary (Bernoulli) dropout \cite{srivastava2014}, $p(\varepsilon)=p^\varepsilon(1-p)^{1-\varepsilon}$ and for Gaussian dropout \cite{wang2013}, $p(\varepsilon) \sim \mathcal{N}(1,p/(1-p))$.

As a generalization of the dropout technique, \cite{kingma2015} propose to treat the distribution of the layer $O_h$ as a posterior distribution of the weights. In case of Gaussian noise, for example, the distribution of $O_h$ before the activation function would also be Gaussian, with mean $X_hW_h$ and variance $p/(1-p)(X_hW_h)^TX_hW_h$. Alternatively, we can see the initialization of the weights as placing a Gaussian process prior on them, which leads to the Bayesian prior-likelihood decomposition of the covariance function parameters $p(\omega|X,Y)$ (\cite{gal2016}). In particular, the likelihood of the data appears in the posterior through the probability $p(y|x,\omega)$.

\subsection{$\alpha-$ divergence}

The Kullback-Leibler divergence, while the most commonly used distribution similarity measure, belongs to the wider class of \textit{Renyi} or $\alpha$-divergences which are parametrized by a hyperparameter $\alpha$ and have the following form:

\begin{equation}
D_\alpha(p(w)||q(w))=\frac{1}{(\alpha-1)}\log\bigg(\int_S p(w)^\alpha q(w)^{1-\alpha}dw\bigg),
\label{eq:renyi_one}
\end{equation}
where $p(w)$ and $q(w)$ are distributions over a common support $S$.

One reason why Renyi (or $\alpha$) divergences have become increasingly popular are their close connection to the classical measures of similarity between distributions, such as Kullback-Leibler and Hellinger divergences.
Table~\ref{tab:renyi_examples} presents an overview of particular values of the parameter $\alpha$ (\cite{hernandez2016}).

\begin{table}[h]
  \caption{The KL and Hellinger measures as special cases of $\alpha$-divergence.}
  \label{tab:renyi_examples}
  \centering
  \begin{tabular}{lll}
    \toprule
    \cmidrule{1-3}
    $\alpha$ & Form of $D_\alpha(p(w)||q(w))$ & Interpretation  \\
    \midrule
    0     & $-\log\int_S q(w)dw$ & Negative log probability      \\
    0.5     & $-2 \log\int_S\sqrt{q(w)}\sqrt{p(w)}dw$ & $-2\log(1-\text{Hellinger}^2)$  \\
    1 & $\int_S p(w)\log\frac{p(w)}{q(w)}dw$  & Kullback-Leibler     \\
    \bottomrule
  \end{tabular}
\end{table}

In addition, $\alpha$-divergences are connected to the $\chi^2$ and total variation divergences (\cite{van2014}).
Using continuity properties, the definition can be extended to any real-valued $\alpha$, giving better flexibility during the training process.

Lastly, notice how Equation~\ref{eq:renyi_one} can be rewritten as
\begin{equation}
D_\alpha(p(w)||q(w))=\frac{1}{(\alpha-1)}\log\mathbb{E}_{q(w)}\bigg[\bigg(\frac{p(w)}{q(w)}\bigg)^\alpha\bigg],
\label{eq:renyi_two}
\end{equation}
and $q(w)=0 \rightarrow p(w)=0\;\forall w \in S$.\\
This form of the $\alpha$-divergence turns out to be more useful if we need to compute mini-batch gradient approximations via Monte-Carlo integration.

\subsection{Variational inference}

In the case when the original problem is intractable, Bayesian methods allow for approximate inference to be performed. Consider a collection of observations $x$ and latent variables $h$. We are then interested in finding the form of $p(h|x)\propto p(x|h)p(h)$, the posterior distribution of the latent variables. However, the exact posterior is intractable, which is why we approximate it with a simpler distribution $q_\phi(h|x)$. The task then becomes to maximize the ELBO defined by:
\begin{equation}
\mathcal{L}=\log p_\theta(x)-D_{KL}(q(h|x)||p(h|x)),
\label{eq:old_ELBO}
\end{equation}
where $D_{KL}$ is the Kullback-Leibler divergence. Because the term $\log p_\theta(x)$ is constant, maximizing the ELBO is equivalent to minimizing the KL divergence. As $D_{KL}(q||p)\rightarrow 0$, the approximate posterior $q_\phi(h|x)$ approaches the true intractable posterior $p_\theta(h|x)$.\\
In the context of this paper, we take the divergence between the prior and the approximate posterior, seeing the network weights as latent variables. The approximate posterior $q(w)\equiv q(w|x,y)$ for the labeled tuples $(x,y)$ of the dataset.

\subsection{Reparametrization trick}

In order to train a neural architecture, it is essential to have provide the optimizer with a differentiable loss function. Stochastic optimization will then be performed using automatically computed gradients with respect to the model parameters. However, when using a Monte Carlo integral approximation to the parameter gradient at a given point, we need to take care of the sampling process.For example, we might be interested in approximating the posterior on the model weights to be a Gaussian $q(w_{ij})=\mathcal{N}(\mu_{ij},\sigma^2_{ij})$. However, if we approximate the gradients using mini-batch, it is important to take the sampling operator out of the network by using the location-scale family representation of the Gaussian: $\mathcal{N}(\mu_{ij},\sigma^2_{ij}) \stackrel{d}{=} \mu_{ij}+\sigma_{ij}\mathcal{N}(0,1)$. This allows to easily take the gradient (i.e., the derivatives with respect to the model parameters $\partial q(w_{ij})/\partial \mu_{ij}$ and $\partial q(w_{ij})/\partial \sigma^2_{ij}$) of the approximate posterior.

\section{Related Work}

Our work extends the variational dropout technique originally proposed in \cite{kingma2015}. It was suggested that applying a multiplicative random noise to inputs of a fully-connected neural network layer is equivalent to computing the posterior distribution of the pre-activation outputs (\cite{kingma2015}). Using the reparametrization of the weights $w$ as $w_i=S_i\theta_i$, where $\theta_i$ is a deterministic weight value and $S_i$ is a random variable determining the scale of the $i^{th}$ weight. As described in \cite{kingma2015}, it is possible to split the posterior distribution over the weights $q_\phi(W)$ into an additive (i.e., location) parameter $\theta$ and a multiplicative (i.e., scale) parameter $a$. Formally, $\phi=\{\theta\} \cup \{a\}$, using set notation.\\

The above decomposition allows us to apply variational approximation techniques by simultaneously maximizing the expected log-likelihood of the parameters and minimizing the divergence between the weights posterior and prior parametrized by $a$.\\
For instance, weights initialization is equivalent to assigning a prior over the weight matrix. The training process then jointly maximizes the expected log-likelihood of the network parameters $\mathbb{E}_{q_a(w)}[\ell_D(\theta)]$ and minimizes the divergence between the prior distribution $p(w)$ and the posterior $q(w)$ on the weights.

The objective function can therefore be bounded by the variational lower bound on the expected log likelihood defined in Equation~\ref{eq:old_ELBO}:
\begin{equation}
\mathcal{L}=\mathbb{E}_{q_a(w)}[\ell_{D}(\theta)]-D_{KL}(q_a(w)||p(w))
\label{eq:old_ELBO}
\end{equation}

Being the relative generalization of Shannon's entropy\cite{lin1991}, KL is by far the most widely known divergence measure. It is very closely related to maximum likelihood estimation since minimizing the KL divergence is equivalent to maximizing the likelihood of the target parameters given the observed data \cite{murphy2012}. Below we detail some essential properties of Kullback-Leibler divergence:
\begin{itemize}
\item $D_{KL}(p||q)\geq 0$ by Gibbs' inequality;
\item $D_{KL}(p_1p_2||q_1q_2)=D_{KL}(p_1||q_1)+D_{KL}(p_2||q_2)$ for $p_1 \bot p_2$ and $q_1 \bot q_2$;
\item $D_{KL}(p||q) \neq D_{KL}(q||p)$ unless $p=q$. Kullback-Leibler hence is not a distance because it is not symmetric. It also does \textit{not} satisfy the triangle inequality.
\end{itemize}

Several recent work proposed replacing the KL divergence in variational inference, as KL highly underestimates model uncertainty \cite{Li17,hernandez2016,li2016renyi}. In this work, we show that by considering $\alpha$ as a hyperparameter and plugging $\alpha$ values at training time, we can still achieve better model performance as $\alpha$ draws to 1, for which KL divergence is the special case.

\section{Divergence metrics in Variational Dropout}

In order to extend the variational lower bound derived by \cite{kingma2015}, we generalize the expression to handle an $\alpha$-divergence parametrized by the hyperparameter $\alpha$. Although the $\alpha$ parameter can theoretically take any real value, it's domain is constrained to positive reals (i.e., $0 \leq \alpha \leq \infty$) over practical concerns. The extension of the divergence to $\alpha=\infty$ is done via limits properties and yields the log supremum of the ratio of both distributions (i.e., $D_\infty(p(w)||q(w))=\log \sup_w \frac{p(w)}{q(w)}$). The variational lower bound is therefore given by, where equation ~\ref{eq:old_ELBO} can be seen as a special case of equation ~\ref{eq:new_ELBO} when $\alpha\rightarrow 1$.

\begin{equation}
\mathcal{L}_\alpha=\mathbb{E}_{q_a(w)}[\ell_{D}(\theta)]-D_{\alpha}(q_a(w)||p(w)).
    \label{eq:new_ELBO}
\end{equation}

Similar to \cite{kingma2015}, we start by decomposing the weights into a magnitude and sign random variables:

\begin{equation}
\begin{split}
    & w_i=s_i|w_i|,\\
    & p(s_i)=0.5^{(s_i+1)/2}0.5^{1-(s_i+1)/2}=0.5,\; s_i \in \{-1,1\},\\
    & p(\log(|w_i|)) \propto c.
    \label{eq:weight_priors}
\end{split}
\end{equation}

The weight magnitude $|w_i|$ receives a log-uniform prior (a prior uniform on the log-scale) to make calculations work out nicely and avoiding further approximations. However, this results in an improper posterior. No practical issues arise due to the fact that we are numerically approximating the posterior up to a location and scaling factor.

We can then use the following property of $\alpha$-divergences:

\begin{equation}
D_\alpha[q(s_i)q(\log|w_i|)||p(s_i)p(\log|w_i|)]=D_\alpha[q(s_i)||p(s_i)]+D_\alpha[q(\log|w_i|)||p(\log|w_i|)]
\end{equation}

This allows us to deal separately with the divergence between the weights' posterior and prior over the sign and the magnitude. We can decompose the above expression further:

\begin{equation}
\begin{split}
     D_\alpha[q(s_i)q(\log|w_i|)||p(s_i)p(\log|w_i|)]=&D_\alpha[q(s_i)||p(s_i)]+D_\alpha[q(\log|w_i|)||p(\log|w_i|)]\\
     =&-\log0.5+\frac{1}{\alpha-1}\log(Q(0)^\alpha+(1-Q(0))^\alpha)\\
     +&D_\alpha[q(\log|w_i|)||p(\log|w_i|)].
    \label{eq:alpha_1}
\end{split}
\end{equation}
Here, $Q(0)$ is the cumulative posterior distribution function of the weight $w_i$ evaluated at 0, which is equivalent to $P(w_i\leq 0)=P(w_i < 0)$, the probability to observe a negative weight. The last equality is allowed by the assumption that the distribution is continuous, i.e. $P(w_i=0)=0$. The term $1-Q(0)$ corresponds to $P(w_i\geq 0)=P(w_i>0)$, the probability that the weight is positive.

We can compute the second term separately by using the definition of $\alpha$-divergence once again:
\begin{equation}
    \begin{split}
        D_\alpha[q(\log|w_i|)||p(\log|w_i|)]=&\frac{1}{\alpha-1}\log\bigintsss_S\frac{p(\log|w_i|)^\alpha}{q(\log|w_i|)^\alpha}q(\log|w_i|)dw_i\\
        \propto & \frac{1}{\alpha-1}\log\bigintsss_S\frac{c^\alpha}{q(\log|w_i|)^\alpha}q(\log|w_i|)dw_i\\
        =& \frac{\alpha}{\alpha-1}\log c+ \frac{1}{\alpha-1}\log\bigintsss_S\frac{1}{q(\log|w_i|)^\alpha}q(\log|w_i|)dw_i\\
        =& \frac{\alpha}{\alpha-1}\log c+ \frac{1}{\alpha-1}\log\bigintsss_S q(\log|w_i|)^{1-\alpha}dw_i\\
    \end{split}
\end{equation}

We reparametrize $\log |w_i|=|w_i|$, which introduces the term $\frac{1}{|w_i|}$ through a Jacobian transformation. We then obtain the following expression:

\begin{equation}
    \begin{split}
        D_\alpha[q(\log|w_i|)||p(\log|w_i|)]=&\frac{\alpha}{\alpha-1}\log c+ \frac{1}{\alpha-1}\log\bigintsss_S q(\log|w_i|)^{1-\alpha}dw_i\\
        =&\frac{\alpha}{\alpha-1}\log c+ \frac{1}{\alpha-1}\log\bigintsss_S q(|w_i|)^{1-\alpha}\frac{1}{|w_i|}dw_i\\
        =&\frac{\alpha}{\alpha-1}\log c+ \frac{1}{\alpha-1}\log\mathbb{E}_{q(|w_i|)}\bigg[ \frac{q(w_i)^{-\alpha}}{|w_i|}\bigg]\\
        =&\frac{\alpha}{\alpha-1}\log c+ \frac{1}{\alpha-1}\log\mathbb{E}_{q(w_i)}\bigg[\bigg(\frac{q(w_i)}{(1-Q(0))}\bigg)^{-\alpha}\frac{1}{|w_i|}+\bigg(\frac{q(w_i)}{Q(0)}\bigg)^{-\alpha}\frac{1}{|w_i|}\bigg],\\
    \end{split}
\end{equation}

where the term $\frac{q(w_i)}{(1-Q(0))}$ truncates the posterior to the negative part of the real line and the term $\frac{q(w_i)}{Q(0)}$ to the positive, respectively. Combining both terms together we get the expression for the negative $\alpha$-divergence between the prior and posterior of the weight $w_i$:

\begin{equation}
    \begin{split}
        -D_\alpha[q(s_i)q(\log|w_i|)||p(s_i)p(\log|w_i|)]=&\log 0.5 - \frac{\alpha}{\alpha-1}\log c-\frac{1}{\alpha-1}\log \mathbb{E}_{q(\varepsilon_i)}\bigg[\frac{q(\varepsilon_i)^{-\alpha}}{|\varepsilon_i|}\bigg],
        \label{eq:final_new_divergence}
    \end{split}
\end{equation}

where we use the reparametrization $w_i=\theta_i\varepsilon_i,\; \partial w_i/\theta_i=\partial \varepsilon_i$ and all terms involving $\theta_i$ cancel out. Now, recall that $\varepsilon_i \sim q(\varepsilon_i)$. For the case when $q(\varepsilon_i)\sim \mathcal{N}(1,p/(1-p))$, we can produce an interpolating polynomial of the third order for various values of the parameter $\alpha$.\\
Due to the fact that the distribution becomes intractable, we resort to Monte Carlo integration to approximate the expectation of the posterior.

In practice, we can use the following approximation to the likelihood of $N$ points using a mini-batch of $M$ points $X$ and respective labels $Y$:

\begin{equation}
\mathbb{E}_{q_\phi(w)}[\log p(Y|X,w)] \approx \frac{N}{M}\sum_{i=1}^M\log p(y_i|x_i,w),
\label{eq:minibatch}
\end{equation}
where $(x_i,y_i)$ is one data point of the mini-batch dataset and $\phi$ is the set of trainable parameters of the network. The gradient can then be computed by interchanging the summation and gradient $\nabla_\phi$. For the special case when the weight posterior is a Gaussian distribution, we pre-compute a 3$^{rd}$ order polynomial interpolating the $\alpha$-divergence and avoid directly using Equation~\ref{eq:minibatch} during training time. If we call the term $p/(1-p)$ as $a$, the variational lower bound becomes differentiable in $a$ and the gradient can flow through.\\

For each value of $\alpha$ examined, $N=100,000$ points were first sampled from a $\mathcal{N}(1,a)$ distribution. The expectation term was then computed by averaging over the draws. A third order polynomial was then fitted for each value of $\alpha$ and used in all further computations involving the $\alpha$-divergence. Figure~\ref{fig:alpha_poly} shows the simulated Monte-Carlo points, as well as the interpolating degree 3 polynomials for $\alpha=0.1,0.95$ and $10$. The shape of the curve $D_{KL}\equiv D_{1}$ is similar to the one obtained by \cite{kingma2015}. We see that $-D_{10}$ is similar in shape to $-D_1$. Since the constant $c$ can be any finite real number due to an improper prior, it's role becomes to shift the curve up or down. We can thus either use it to make the curve pass by the point $(1,0)$ or take $c=1$ to get rid of the term completely. We chose the second option due to simplicity.

It is worth mentioning that when a particular family of prior (e.g., log-uniform) is fixed, the $\alpha$-divergence can indeed be calculated exactly by simulating draws from the posterior, the prior and subsequently appealing to the law of large numbers to approximate the expectation. However, the methodology proposed by \cite{kingma2015} works for most of the reasonable prior-posterior combinations.

\section{Experimental Results}

We compare the significance of using $\alpha$-divergence with variational dropout on a set of tasks based on the MNIST dataset. Two variational dropout constructions were considered from \cite{kingma2015}: variational A (with correlated weight noise) and variational B (with independent weight noise). Both networks were trained with the ELBO (Equation \ref{eq:new_ELBO}) as loss function. Early stopping was used to achieve a reasonable training time. All figures were obtained by taking the average of test set classification error and accuracy evaluated on the MNIST dataset over 5 runs.

\begin{figure}[h]
    \centering
    \subfloat[Correlated noise dropout.\label{fig:varA_size}]{\includegraphics[width=0.4\textwidth]{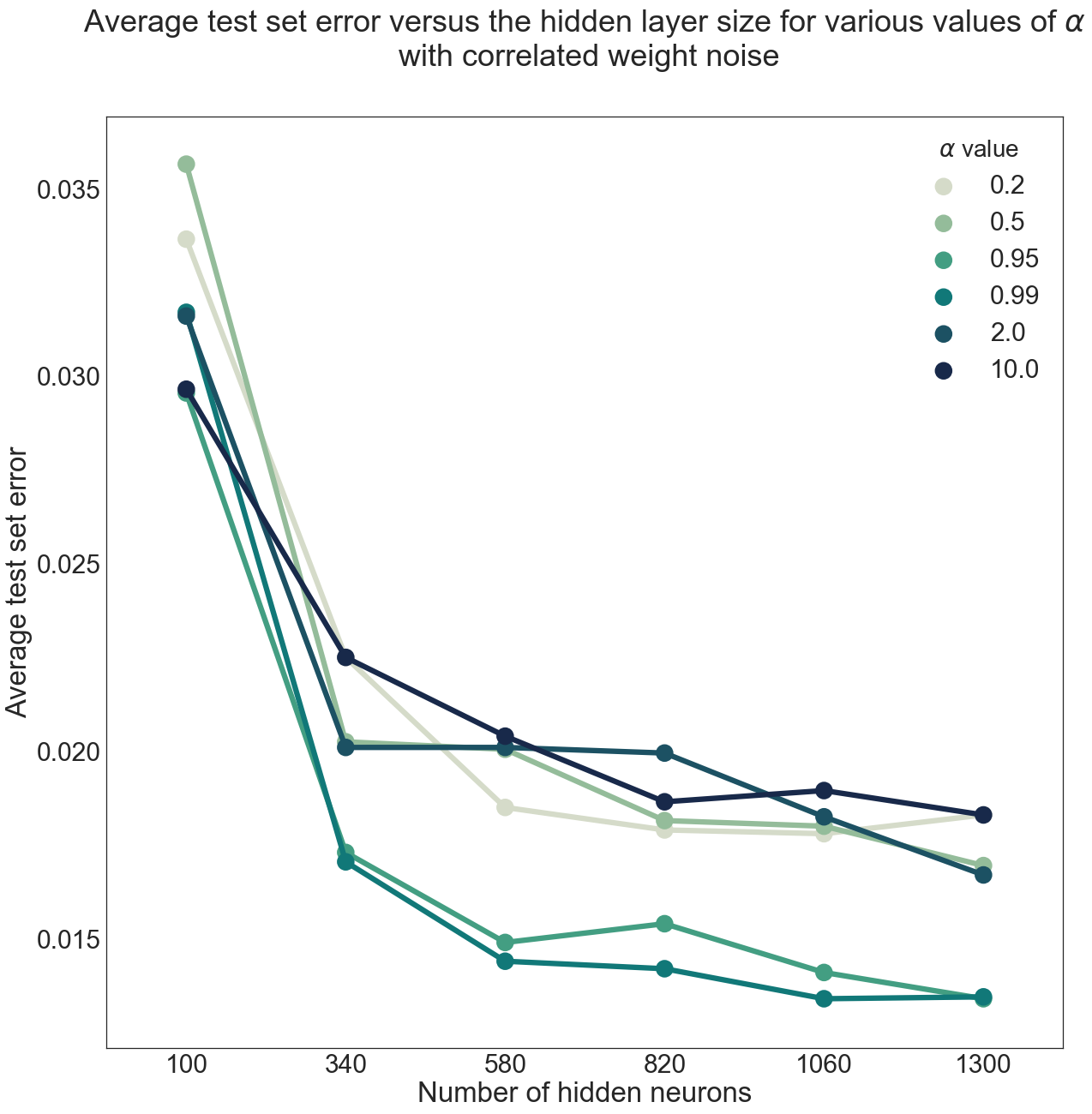} }
\hfill
    \subfloat[Uncorrelated noise dropout.\label{fig:varB_size}]{\includegraphics[width=0.4\textwidth]{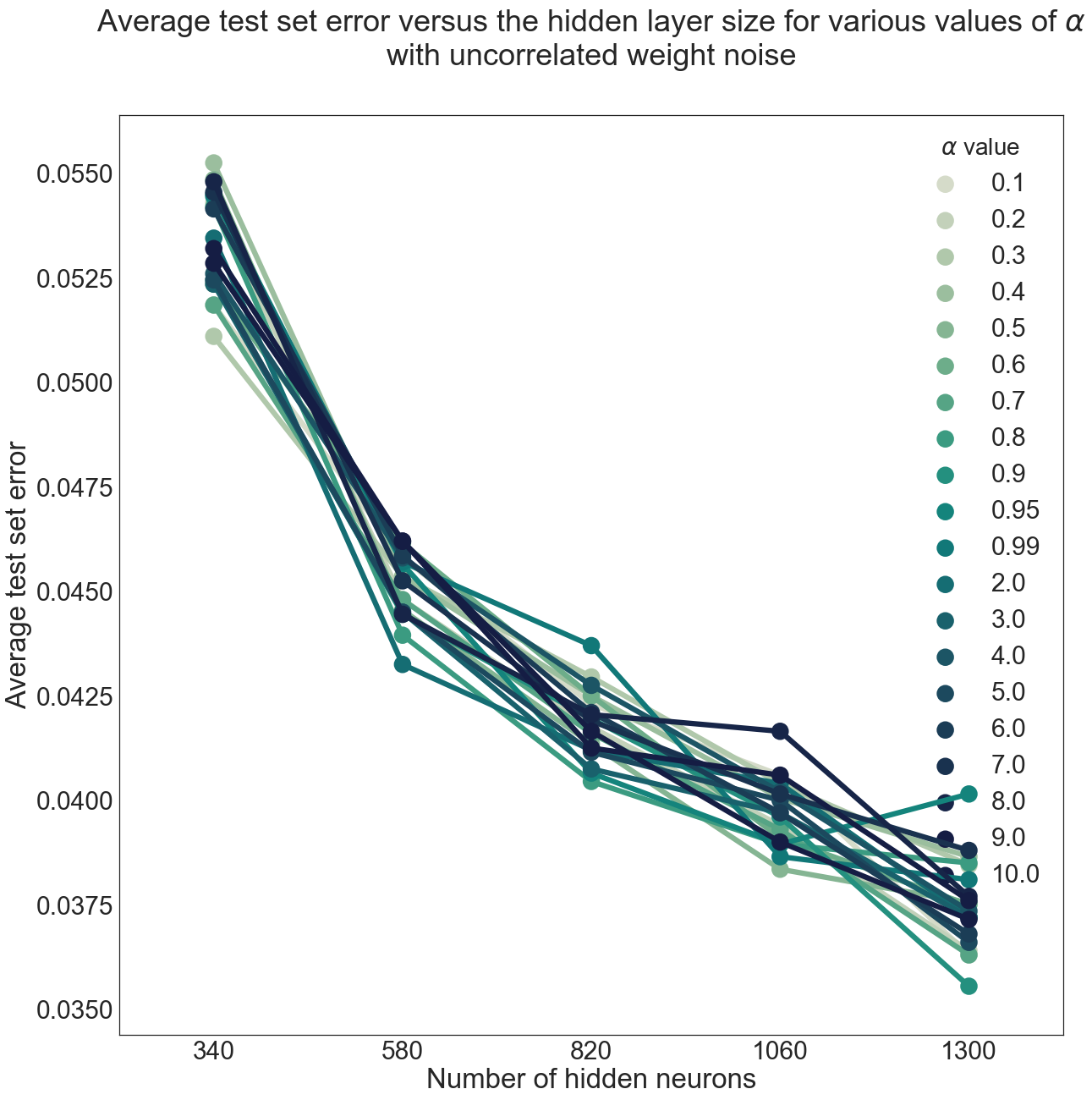} }
    \caption{Average test set errors as functions of either the hidden layer size or the divergence parameter $\alpha$ for variational dropout with \textbf{(a)} correlated weight noise (varA) and \textbf{(b)} uncorrelated weight noise (varB).}
    \label{fig:results_error}
\end{figure}

As shown in Figure~\ref{fig:varA_size}, training the variational A dropout network with an $\alpha$-divergence ELBO with $\alpha \rightarrow 1$ yields the lowest test set error. As the size (i.e., number of hidden neurons) of the network increases, taking a divergence close to Kullback-Leibler gives a lower average error than any other $\alpha$-divergence. Figure~\ref{fig:varB_size}
shows that the variational B architecture with uncorrelated noise does not exhibit the same behaviour; all values of $\alpha$ perform equally well at test time.

\begin{figure}[h]
    \centering
    \subfloat[Average test set classification errors on the MNIST dataset as a function of the hyperparameter $\alpha$ computed for multiple sizes of the hidden layer with correlated noise.\label{fig:varA_alpha}]{\includegraphics[width=0.4\textwidth]{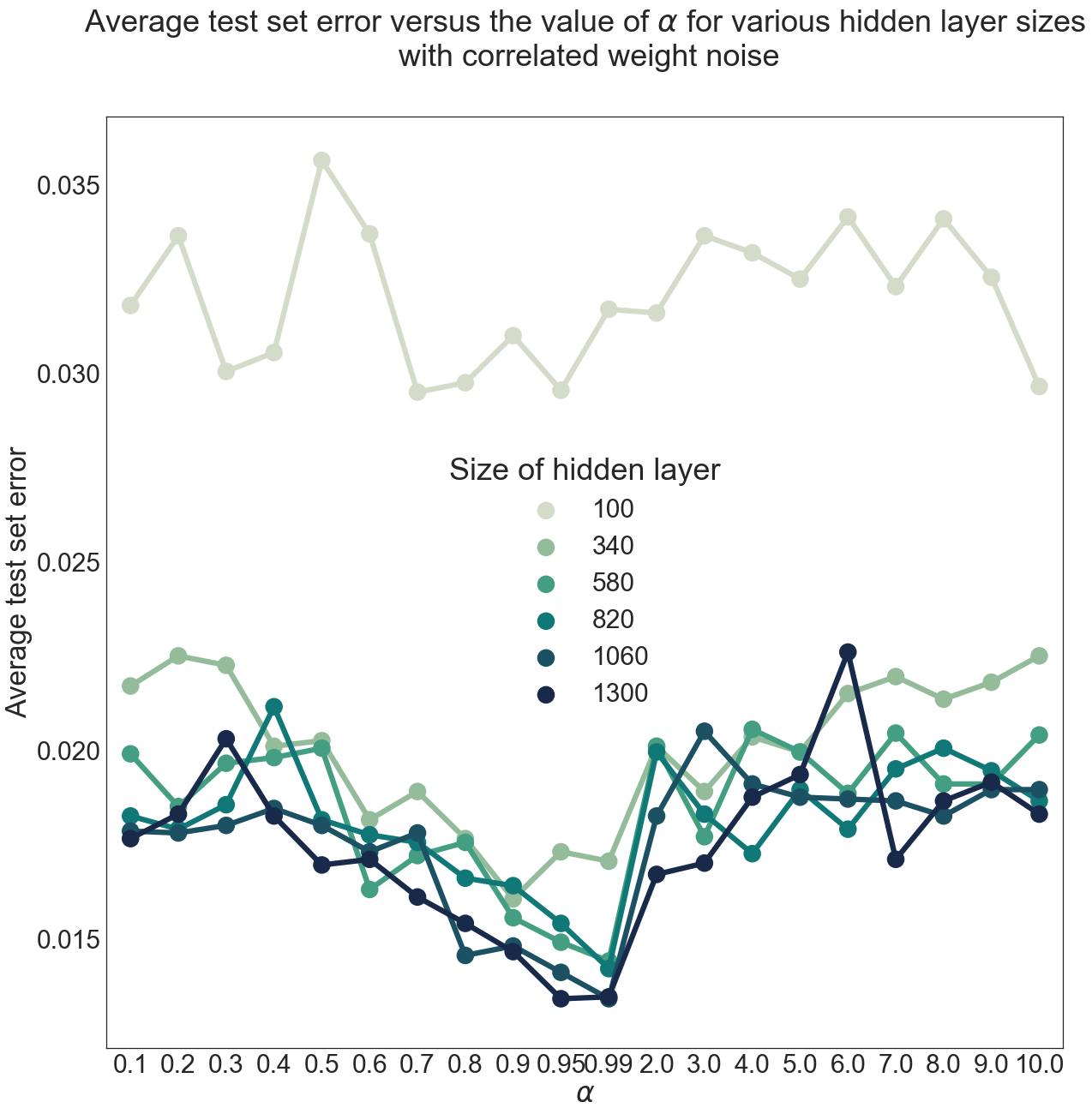} }
    \hfill
    \subfloat[Third order polynomial approximating $-D_\alpha$ for $\alpha=0.1,0.95,10$. \label{fig:alpha_poly}]{\includegraphics[width=0.39\textwidth]{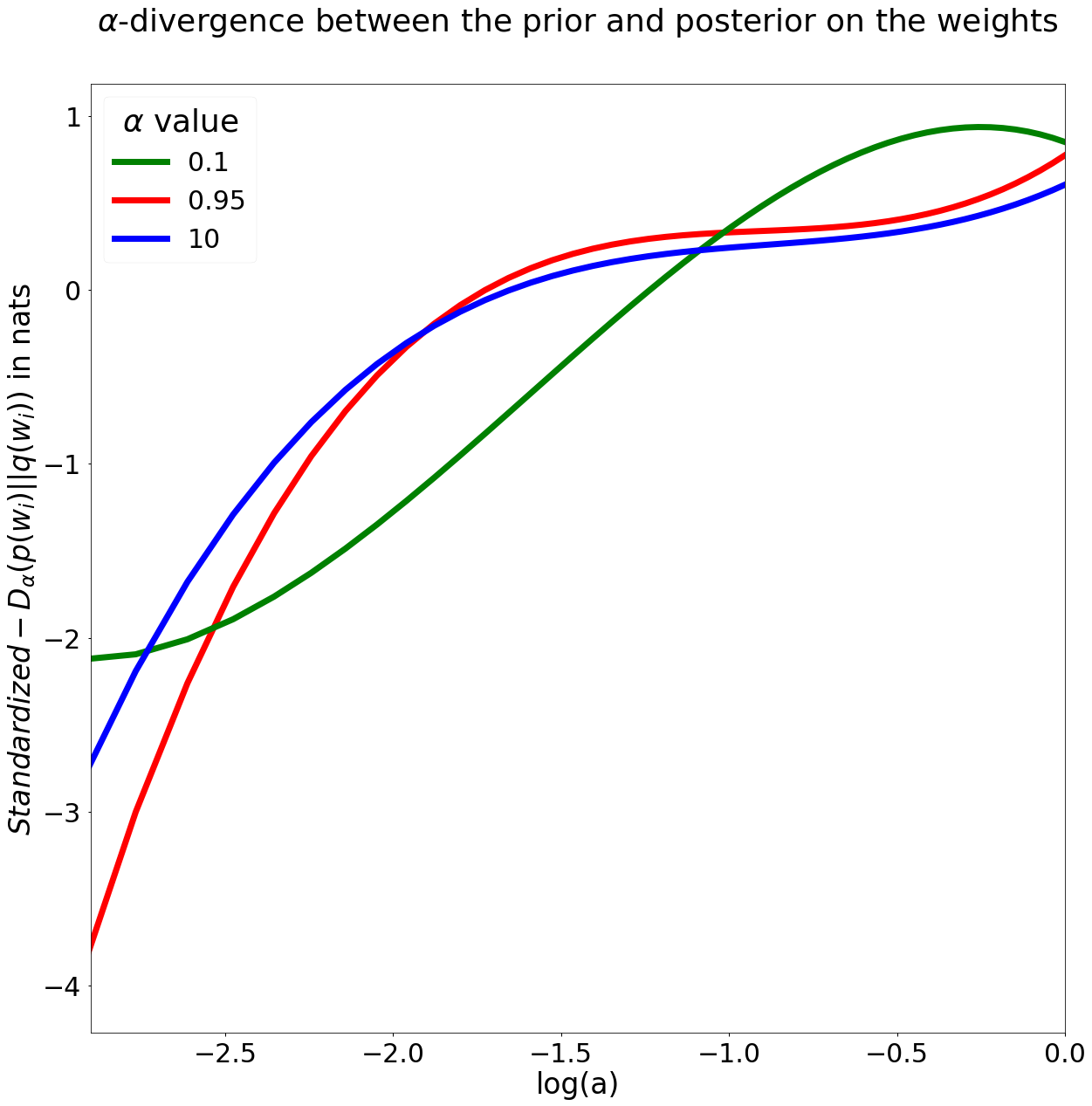}}
    \caption{Classification errors and best fit polynomial for various values of $\alpha$.}
    \label{fig:results_class_poly}
\end{figure}

Figure~\ref{fig:varA_alpha} presents a rearrangement of the test set errors from variational A dropout. It can be seen that the minimum of the function is located at the value $\alpha=1$ which corresponds to the KL divergence. In other words, the variational A dropout with yield the lowest test set error when the variational lower bound is a function of the Kullback-Leibler divergence. Figure~\ref{fig:alpha_poly} shows the standardized $-D_\alpha(p||q)$ for three values of the parameter $\alpha$. We see that as the shape parameter $\alpha$ approaches 1, $-D_\alpha$-divergence converges to the negative KL divergence.

\begin{figure}[h]
    \centering
    \subfloat[Average test set classification accuracy on the MNIST dataset as a function of $\alpha$ versus epochs.\label{fig:varA_accuracy_vs_epochs}]{\includegraphics[width=0.4\textwidth]{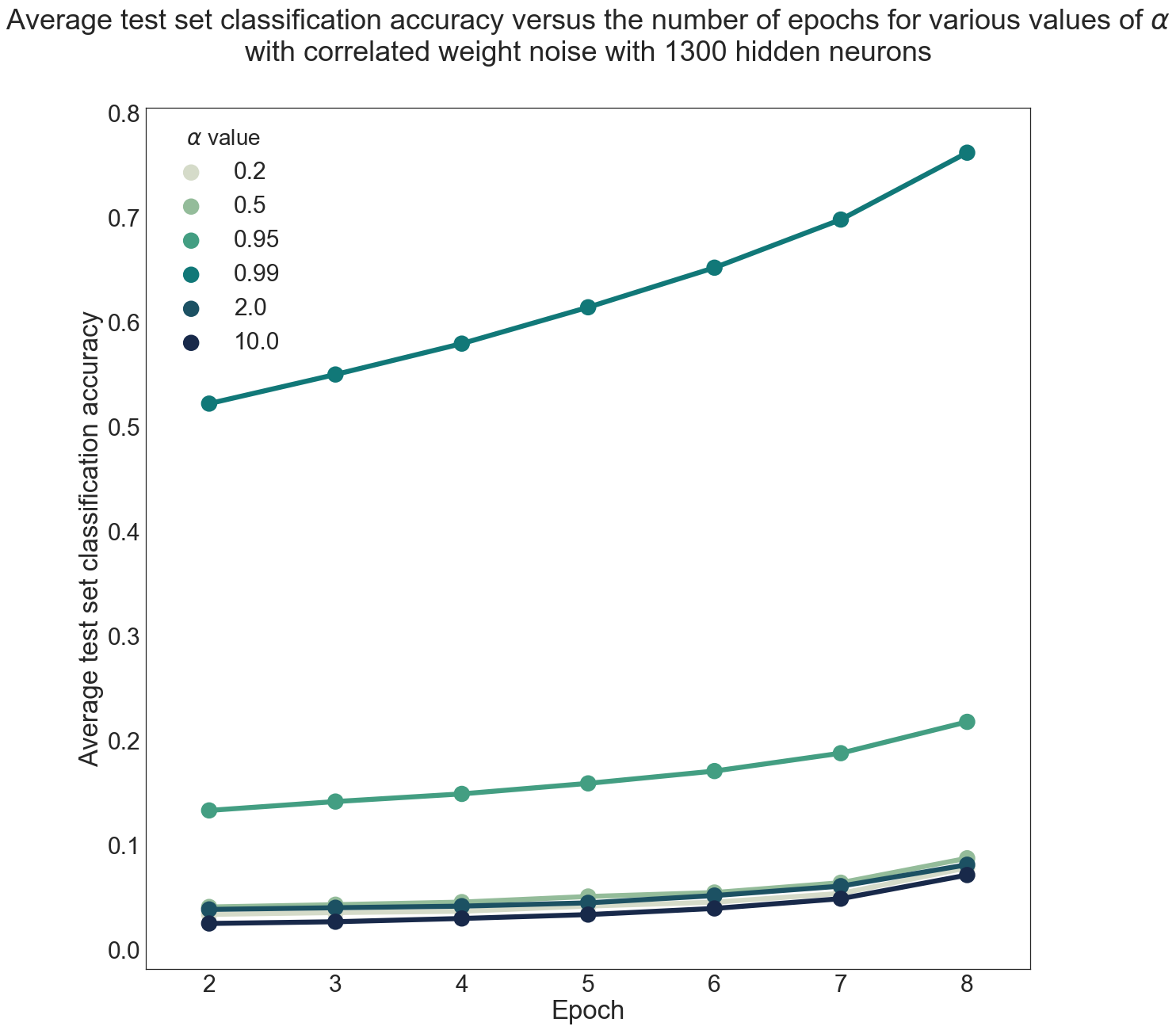} }
    \hfill
    \subfloat[Average test set classification accuracy on the MNIST dataset as a function of $\alpha$ versus hidden layer size. \label{fig:varA_accuracy_vs_size}]{\includegraphics[width=0.4\textwidth]{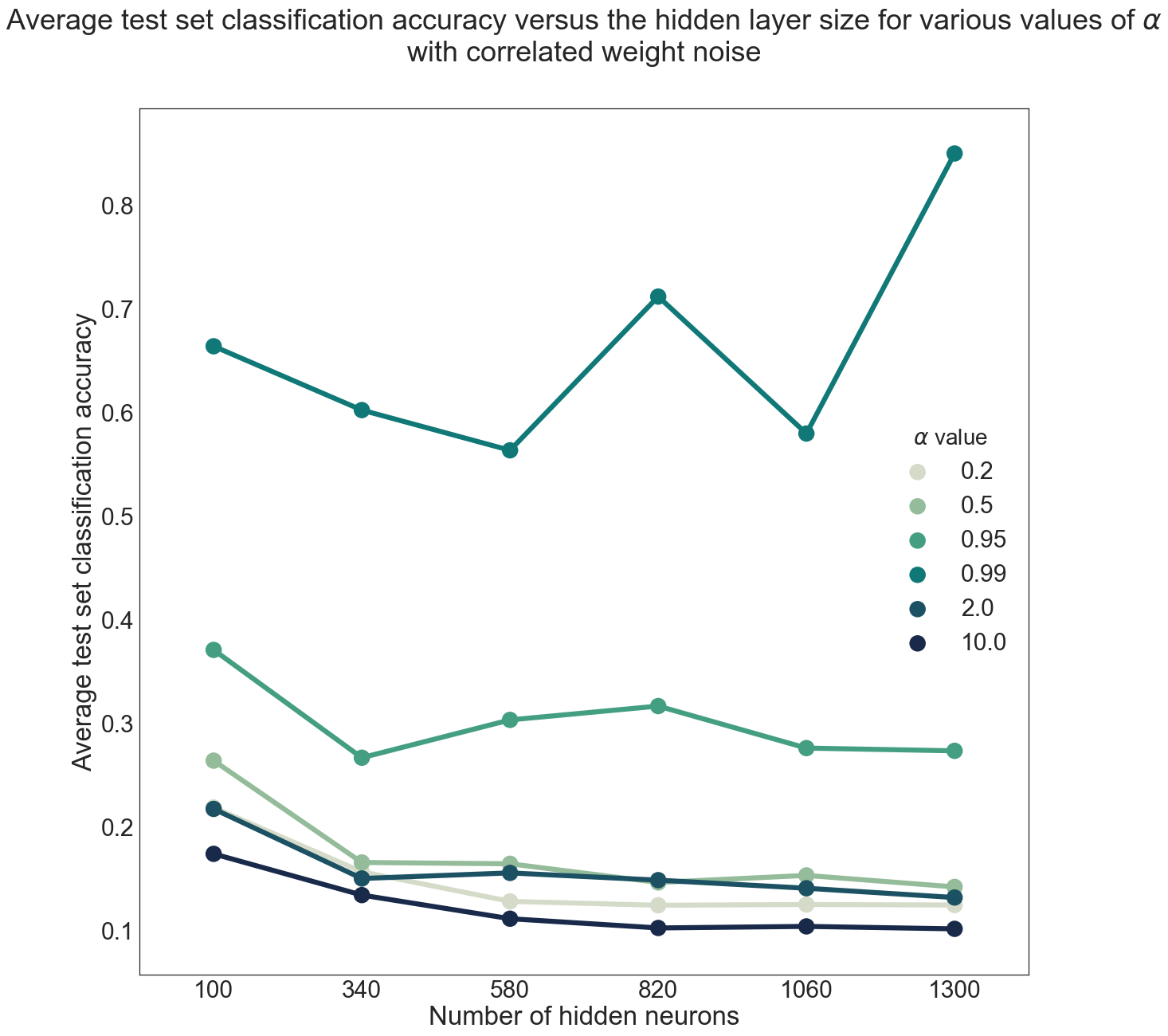} }
    
    \caption{Classification accuracy of variational dropout with correlated noise.}%
    \label{fig:results_accuracy}%
\end{figure}

Figure~\ref{fig:varA_accuracy_vs_epochs} shows the test set classification accuracy as a function of epochs. We see that $\alpha=0.99$ yields the best performance, followed by $\alpha=0.95$. If we refer to Figure~\ref{fig:varA_accuracy_vs_size}, a similar pattern can be observed: when the value of $\alpha$ approaches 1 from below, the classification accuracy improves drastically.

The results suggest that if we represent the KL divergence as a special case of $\alpha$-divergence family parametrized by a shape variable, then taking $\alpha=1$ will yield the lowest error or equivalently the highest classification accuracy for variational dropout with correlated noise.

\section{Discussion and Conclusion}

In this work, we analyzed the extension of variational dropout with alpha-divergence minimization. Variational dropout \cite{kingma2015} relies on a Bayesian treatment of the weights and minimizes the Kullback-Leibler divergence between the prior and approximate posterior of the weights. In the case of correlated weight noise, the weight $w_i$ can be decomposed into a deterministic parameter $\theta_i$ and a random scale variable $S_i$. The variational lower bound is then used to maximize the likelihood of the observations. We presented a generalization of the variational dropout method to the $\alpha$-divergence family. The divergence term in the ELBO expression is still intractable. We therefore resort to third order polynomial interpolation of the $\alpha$-divergence as a function of the dropout rate $a$ in the case of Gaussian dropout. We provide empirical evidence from the MNIST dataset which suggested that, when using alpha-divergences, the Kullback-Leibler divergence ($D_\alpha \rightarrow D_{KL},\; \; \alpha \rightarrow 1$) yields the lowest average test set classification error for variational dropout with correlated weight noise. This provides an argument in favor of using the KL divergence in variational dropout as opposed to generalized $\alpha$-divergences.\\

In future work, it might be interesting to adopt a fully Bayesian treatment of the $\alpha$-divergence. For instance, instead of manually tweaking $\alpha$, we would like the model to capture both the knowledge that KL performs better than other $\alpha$-divergences in variational dropout along with model uncertainty. We could introduce a prior $p_\chi(\alpha)$ defined on positive reals with the parameters $\chi$ chosen such that the prior has a peak at $\alpha=1$. 


\bibliography{main}
\bibliographystyle{iclr2018_conference}

\end{document}